\documentclass{article}

\usepackage{microtype}
\usepackage{graphicx}
\usepackage{subfigure}
\usepackage{booktabs} 

\usepackage{hyperref}
\hypersetup{
    pdftitle={NeuRIPS CCAI Workshop 2020 - MuSTAS From Talk to Action with Accountability}
}

\usepackage[accepted]{tccmlicml2021}

\usepackage{multirow}

\icmltitlerunning{Monitoring Policy Makers with Deep Neural Networks and Topic Modelling}

\begin{document}

\twocolumn[
    \icmltitle{From Talk to Action with Accountability: Monitoring the Public Discussion of Policy Makers with Deep Neural Networks and Topic Modelling\\
    \textit{Proposals Track}}
    
    
    
    \icmlsetsymbol{equal}{*}
    
    \begin{icmlauthorlist}
    \icmlauthor{Hätönen, Vili}{equal,Em}
    \icmlauthor{Melzer, Fiona}{equal,UoE}
    
    \end{icmlauthorlist}
    
    \icmlaffiliation{UoE}{School of Psychology, Philosophy and Language Science, University of Edinburgh, Edinburgh, Scotland}
    \icmlaffiliation{Em}{Emblica Oy, Helsinki, Finland}
    
    \icmlcorrespondingauthor{Fiona Melzer}{f.melzer@sms.ed.ac.uk}
    \icmlcorrespondingauthor{Vili Hätönen}{vili@emblica.com}
    
    \icmlkeywords{Machine Learning, ICML, NLP, Topic Modelling, Transparency, Accountability}
    
    \vskip 0.3in
]



\printAffiliationsAndNotice{\icmlEqualContribution} 

\begin{abstract}
 Decades of research on climate have provided a consensus that human activity has changed the climate and we are currently heading into a climate crisis. 
 While public discussion and research efforts on climate change mitigation have increased, 
 potential solutions need to not only be discussed but also effectively deployed.
 For preventing mismanagement and holding policy makers accountable, transparency and degree of information about government processes have been shown to be crucial.
 However, currently the quantity of information about climate change discussions and the range of sources make it increasingly difficult for the public and civil society to maintain an overview to hold politicians accountable. 
 In response, we propose a multi-source topic aggregation system (MuSTAS) which processes policy makers speech and rhetoric from several publicly available sources into an easily digestible topic summary. MuSTAS uses novel multi-source hybrid latent Dirichlet allocation to model topics from a variety of documents. 
 This topic digest will serve the general public and civil society in assessing where, how, and when politicians talk about climate and climate policies, enabling them to hold politicians accountable for their actions to mitigate climate change and lack thereof.
 
\end{abstract}

\section{Introduction}
\label{Introduction}

The consensus that human activity caused the climate crisis \cite{cook2016consensus} has led to the development of many tools and agreements, designed to support climate change mitigation efforts. Most notably the Paris Agreement, adopted by 197 countries that aims to keep global warming below 2°C \cite{paris}. 
However, research efforts to mitigate the climate crisis are lost without an efficient international adaptation of proposed tools and policies.

Scientists, non-state actors
, and voters increasingly critique their government for insufficient action mitigating climate change \cite{dupont2012insufficient}.
This suggests a gap between promises made by politicians and actual action taken: ambitious promises for climate change mitigation have turned into careless discourse with insufficient measures taken.

Holding politicians accountable for their actions has been shown to be a major factor in preventing mismanagement, political corruption and misalignment of politician’s opinions and the public they are representing \cite{adsera2003you,lyrio2018thirty}. Transparency herein is so crucial, that it has been considered a direct substitute for accountability in the democratic process \cite{lyrio2018thirty}.
Political transparency, unfortunately, has been made increasingly difficult, not due to a lack, but because of the overwhelming quantity of data accessible: The public and civil society organizations, lack the overview of all that politicians are discussing across a multitude of platforms such as interviews, blog posts, parliamentary speeches and social media posts.
Our work therefore, provides a tool for transparency, supporting civil society in efforts to monitor political discourse around climate change and hold policy makers accountable for their promises and claims. 

In Section \ref{section-MuSTAS} we introduce a Multi-Source Topic Aggregation System (MuSTAS) which increases transparency by providing an overview of topics discussed by politicians across a broad range of platforms. Additionally, MuSTAS provides a user interface with topic summaries and tagged source-texts.

In Section \ref{section-ml} we describe the ML foundation for MuSTAS: a novel multi-source hybrid latent Dirichlet allocation model which forms the core of this research proposal. In Section \ref{section-climate} we outline how MuSTAS impacts climate change.

\section{Multi-Source Topic Aggregation System}\label{section-MuSTAS}

To help 
holding policy makers accountable, MuSTAS processes documents (e.g. speech transcripts, tweets, blog posts) from several publicly available sources into a compact report, highlighting the distribution of topics the candidate or party has been discussing during the given time period on the given platforms. 

\begin{figure}[h!]
    \centering
    \includegraphics[width=\columnwidth,trim={0 36 0 0},clip]{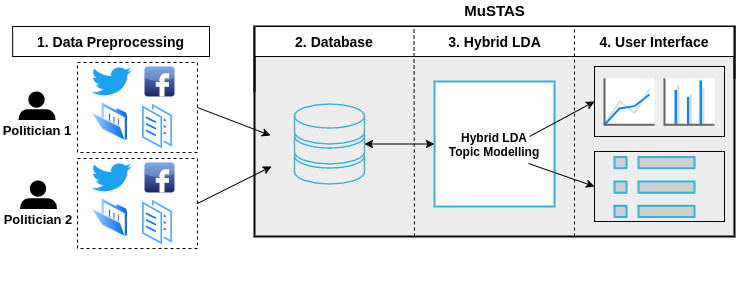}
    \vskip -0.1in
    \caption{Overview of MuSTAS.
    1) Crawling and pre-processing data from Twitter API, Facebook API, blog posts, parliamentary transcripts, etc. produced by the given politician. 
    2) Database for storing the documents and modelling results. Each processed entry includes topic, person, party, timestamp, data type, and source URL, and document metadata.
    3) Topic modelling over all document types using a hybrid latent Dirichlet allocation as described in Section \ref{section-ml}.
    4) Users can access the results through a graphical user interface, which supports different analyses by topic, time, person, platform, and example documents for the given topic.
    }
    \label{fig:MuSTAS}
    \vskip -0.15in
\end{figure}

Through MuSTAS, the public and civil society will have access to infographics that provide answers to questions regarding candidates or groups of candidates, topics discussed, and mediums used, over different time periods (Figure \ref{fig:MuSTAS} step 4).
The questions answered with MusTAS could be e.g. "Do parliamentary discussions address climate change?", "How much does the politician X talk about climate in contrast to economy?", "Does the political discourse differ between platforms meant to address the electorate (such as social media) and political spheres (parliamentary discussions)?", and "Does candidate X talk about different topics prior to the elections in comparison to their time in the government?". 

With answers to these questions voters have better insight what their candidate/party discuss in parliamentary sessions after being elected.
The civil society organisations can assess how much each politician talks about climate in contrast to other topics, which enables efficient targeting of individuals that need convincing on the urgency of climate friendly policies.

Machine learning methods are crucial for this analysis, as hundreds of politicians produce documents every day, and expensive human-conducted surveys are not made regularly. 
To enable the system to read and analyze the for-humans-intractable amount of data and update the analysis by doing online inference on daily basis, topic modelling has to be outsourced to a ML model. 

\section{Topic Modelling: Multi-Source Hybrid LDA} \label{section-ml}

Topic modelling is a widely used approach to describe the content of text documents through a limited number of topics \cite{yi_comparative_2009}, and used previously also on isolated legislative speech \cite{glavas_computational_2019}. 
The topics are seen as latent variables defining a probability distribution for the vocabulary of words in the document.
The probability distributions of topics are modelled with a probabilistic method such as latent Dirichlet allocation (LDA) \cite{blei_latent_2003}, which has been implemented in many different use cases \cite{heidenreich_media_2019, rehurek_software_2010, tong_text_2016, uys_leveraging_2008, yi_comparative_2009} to model topics that are present in a document.

In order to build a holistic understanding which topics a politician discusses, one needs to gather data from many different media, see phase 1 of Figure \ref{fig:MuSTAS}. 
This poses a novel challenge, since methods which successfully model topics in tweets and other short documents, such as Biterm Topic Model \cite{chen_user_2015}, Supervised LDA \cite{jonsson_evaluation_2015, resnik_beyond_2015}, or LDA-U \cite{jonsson_evaluation_2015}, differ from the models designed for longer documents and large corpora \cite{rehurek_software_2010}.

We propose a multi-source approach, where all documents are processed to paragraphs $p$ of similar length. For example, a tweet could equal to one paragraph while a speech might be split into several. Topic modelling is then performed on these paragraphs individually, and a document's topic distribution is attained by aggregating the distributions of its paragraphs.

Different types of documents might radically vary in vocabulary and style. 
To mitigate this, we propose to use a hybrid LDA \cite{moody_mixing_2016} approach, where the easily interpretable LDA has been made more robust by utilizing word embeddings \cite{mikolov_distributed_2013} provided by transformer models such as BERT \cite{devlin_bert_2019} or RoBERTa \cite{liu_roberta_2019}. 
The recently observed performance of large transformer models \cite{brown_language_2020} suggests that the word-, document-, and topic embeddings applied in a hybrid LDA \cite{moody_mixing_2016} could be learned from a corpora of various types of documents. 
In climate change mitigation, hybrid LDA has been previously proposed for monitoring climate change technology transfer \cite{kulkarni_using_2020}.

The topics are learned in unsupervised manner, and therefore depend on the available data.
The number of topics, defined by the researcher before training, will affect how granular the found topics will be\footnote{For example, with 20 different topics all climate change related discussion could be part of e.g. two topics, while 200 topics will result much more granular topics and discussion on e.g. land usage or fossil fuels can be expected to belong to different topics.}. The suitable number of topics will be experimented during the research and documented in the findings. 

The topic model will be trained offline before made available in MuSTAS. This ensures that the topic modelling results are comparable over time. Later models trained with updated datasets can be released to MuSTAS alongside the original model.

\subsection{Empirical Evaluation - Modelling the Topics of Finnish Politicians} \label{sub-finland}

We will demonstrate the functionality of multi-source hybrid LDA as part of MuSTAS by implementing the system in the Finnish context.
In the case of Finland, a major portion of public statements and discussions of politicians can be covered by the data sources provided in Table \ref{table-data}. 
All the sources are publicly available on the internet, so they can be programmatically gathered to the MusTAS database (Figure \ref{fig:MuSTAS}, step 1).

\begin{table}[t]
  \caption{Example data sources}
  \label{table-data}
  \vskip 0.1in
  \begin{scriptsize}
  
  \centering
  \begin{tabular}{lll}\toprule

    Name                                & Document type               & URL \\
    \midrule
    \multirow{2}{2cm}{Finnish parliament database}  & \multirow{3}{2.3cm}{MP's speeches, written questions, proposals, law initiatives etc.} & \href{https://www.eduskunta.fi/FI/search/Sivut/Vaskiresults.aspx}{www.eduskunta.fi/FI/search/...}
    \\\vspace{0.5cm}\\ 
    Social media APIs           & \multirow{2}{2.3cm}{Policy makers' social media posts} & e.g. \href{https://developer.twitter.com/en/docs/twitter-api}{developer.twitter.com/...}
    \vspace{0.5cm}\\
    Blogs                       & Blog posts                        & e.g. \href{https://puheenvuoro.uusisuomi.fi/author/paavovayrynen/}{puheenvuoro.uusisuomi.fi/...}
    \\
    \bottomrule
  \end{tabular}
  \end{scriptsize}
  \vskip -0.1cm
\end{table}

The documents related to each decision maker, are retrieved from each source and pre-processed to a common format required by the hybrid LDA. 
Since the majority of documents produced by Finnish politicians are in Finnish, the transformer used should be compatible with the language, for example FinBERT \citep{virtanen_multilingual_2019}.


\section{Climate Impact} \label{section-climate}

MuSTAS provides researchers, civil society, and the general public with an easily interpretable digest of topics which the policy makers have talked about.

To reach the goals of the Paris Agreement and mitigate climate change, politicians need to implement a range of ambitious policies.
Increased accountability of government officials has been shown to be crucial in preventing political mismanagement \citep{adsera2003you}. Because an accountable actor needs to provide information and justification for their decisions \citep{mees2019framework}, the degree of information (transparency) citizens have about their government's actions is one of two major hinges in accountability \citep{adsera2003you}. 

Currently, despite many democratic governments providing greater access to government proceedings, transparency is hindered by the sheer amount of data accessible. Media coverings of political developments are biased toward specific topics \citep{eady2019many}, actors \citep{snyder2010press,shor2015paper,vos2013vertical}, and private political interests \citep{statham1996television}.

The proposed MuSTAS (Section \ref{section-MuSTAS}) overcomes these barriers by jointly analysing a wide range of publicly available data and presenting data analyses in accessible and dynamic form without representation- or political-bias. 
MuSTAS allows the public to monitor political speech across platforms by bringing a concise digest of politician's statements, easily accessible with links to the original documents. Therefore, MuSTAS is a tool to track rhetorical political commitments on climate change, equipping civil society and voters with the means to hold decision makers accountable and incentivize policy makers to follow through with legitimate policies.


The proposed work implements MuSTAS in the Finnish context (Subsection \ref{sub-finland}). 
However, the flexible system allows a for a straight-forward implementation in other countries by simply providing basic information on the national politicians such as social media account IDs, blog URLs or links to speeches and law initiatives.

This way MuSTAS can crawl and process the provided source-documents for topic modelling regardless of the country or other context of the politician. 
For the new context, and possibly language, a new topic model needs to be trained, but the data processing and visualisations of the topic digests can be done using the same code base. 
This paves way for implementing MuSTAS in other countries, providing data for accountability in all parliamentary democracies.

\bibliography{references}
\bibliographystyle{icml2021}



\end{document}